\documentclass[runningheads]{llncs}
\usepackage[T1]{fontenc}
\usepackage{graphicx}
\usepackage{multirow}
\usepackage{amssymb,amsmath}
\usepackage{booktabs}

\begin{document}
\title{PlankFormer: Robust Plankton Instance Segmentation via MAE-Pretrained Vision Transformers and Pseudo Community Image Generation}
\titlerunning{PlankFormer: Robust Plankton Instance Segmentation}
%
\author{Masaharu Miyazaki\inst{1} \and
Yurie Otake\inst{2}\orcidID{0000-0002-7607-4132} \and
Koichi Ito\inst{1}\orcidID{0000-0001-7431-7105} \and
Wataru Makino \inst{3}\orcidID{0000-0003-3240-3763} \and
Jotaro Urabe \inst{3}\orcidID{0000-0001-5111-687X} \and
Takafumi Aoki \inst{1}\orcidID{0000-0001-8308-2416}
}
\authorrunning{M. Miyazaki et al.}
%
\institute{Graduate School of Information Sciences, Tohoku University
6-6-05, Aramaki Aza Aoba, Sendai, 9808579, Japan. \and
The Center for Ecological Research, Kyoto University,\\ 2--509--3, Hirano, Otsu-shi, Shiga-ken, 5202113, Japan. \and
Graduate School of Life Sciences, Tohoku University,\\ 6--3, Aramaki Aza Aoba, Aoba-ku, Sendai-shi, 9808578, Japan.}
\maketitle              
\begin{abstract}

  Plankton monitoring is essential for assessing aquatic ecosystems but is limited by the labor-intensive nature of manual microscopic analysis.
  Automating the segmentation of plankton from crowded images is crucial, however, it faces two major challenges: (i) the scarcity of pixel-level annotated datasets and (ii) the difficulty of distinguishing plankton from debris and overlapping individuals using conventional CNN-based methods.
  To address these issues, we propose PlankFormer, a novel framework for plankton instance segmentation.
  First, to overcome the data shortage, we introduce a method to generate labeled Pseudo Community Images (PCI) by synthesizing individual plankton images onto diverse backgrounds, including those created by generative models.
  Second, we propose a segmentation model utilizing a Vision Transformer (ViT) backbone with a Mask2Former decoder.
  To robustly capture the global structural features of plankton against occlusion and debris, we employ a Masked Autoencoder (MAE) for self-supervised pre-training on unlabeled individual images.
  Experimental results on real-world datasets demonstrate that our method significantly outperforms conventional methods, such as Mask R-CNN, particularly in challenging environments with high debris density.
  We demonstrate that our synthetic training strategy and MAE-based architecture enable high-precision segmentation with requiring less manual annotations for individual plankton images.

\end{abstract}

\section{Introduction}

Aquatic ecosystems play a vital role for the environment and society, such as maintaining water quality, regulating air quality, and supplying food and water.
Regular monitoring of aquatic ecosystems is essential to maintain their health.
Zooplankton, which underpin the aquatic food web, serve as crucial indicators of ecosystem status since their species composition and abundance fluctuate sensitively in response to environmental changes \cite{Plankton}.
Therefore, plankton monitoring is regularly conducted in oceans and lakes to survey species and population counts.
However, conventional monitoring relies on manual analysis by experts using optical microscopes, which is time-consuming and labor-intensive.
Furthermore, although this task requires specialized knowledge and experience, the number of skilled experts is declining, making the shortage of human resources a serious problem. 
Consequently, establishing technology to automatically detect and identify individuals from crowded plankton images captured by optical microscopes is an urgent task for sustainable, high-precision monitoring.

In research on automating plankton monitoring, classification methods that take isolated individual images as input have been primarily investigated \cite{2018-LOM-Luo,Lumini1,Kyathanahally-FM-2021,Ito-ACPR-2023}.
However, images obtained from optical microscopes in actual monitoring are ``crowded images (community images)'' containing multiple individuals.
Therefore, a segmentation process to detect and extract individuals from these community images is required as a pre-processing step for automatic classification.
Deep learning-based approaches, such as Convolutional Neural Networks (CNNs) \cite{DL} and Transformers \cite{Vaswani2017AttentionIA}, have become mainstream for instance segmentation, with many models proposed \cite{He2017MaskR,Cheng2021MaskedattentionMT}.
Since these models achieve high accuracy and can be adapted to various domains through specific training, their application to plankton detection has also been explored \cite{Bergum_2020,Panaotis2022ContentAwareSO}.
However, most existing methods target general objects (e.g., people or cars in natural images) and fail to address plankton-specific challenges.
Plankton sizes vary significantly, ranging from rotifers ($\approx 100~\mu\text{m}$) to copepods ($\approx 1~\text{mm}$).
Moreover, in community images, shape diversity is extremely high due to occlusions by debris or other individuals, as well as appearance changes caused by pose variations.
Since general CNN-based methods rely on local features for region estimation, their detection accuracy degrades significantly under such size variations and high shape diversity, especially in occluded conditions.
To accurately detect plankton susceptible to occlusion and pose changes, utilizing global features of the image is crucial. 
Vision Transformer (ViT) \cite{Dosovitskiy-ICLR-2021} is effective for extracting global image features compared to CNNs.
Therefore, in this paper, we propose {\it PlankFormer}, an image segmentation model employing a ViT encoder and a Mask2Former decoder \cite{Cheng2021MaskedattentionMT}.

On the other hand, training an image segmentation model to detect plankton requires a community image dataset with pixel-level annotations for each individual.
However, no such public dataset currently exists.
Creating a dataset from scratch would require manually annotating every individual in the images, which is impractical.
To address the problem of training data scarcity, we propose a method to automatically generate labeled ``Pseudo Community Images (PCI)'' by synthesizing a small number of labeled individual plankton images onto background images.
In our method, PCI and their corresponding ground truth labels are automatically generated by compositing pixel-level labeled individual images onto various backgrounds.
During synthesis, diverse plankton variations are reproduced by applying random flipping, rotation, and resizing to the individual images.
Furthermore, to improve the domain diversity of PCI, we use not only real background regions extracted from actual community images but also images generated by generative models.
However, since PCI is generated by repeatedly using a small number of individual images, the diversity of individuals may be lower than in actual community images.
To capture the features of plankton with drastic pose changes, the model needs to be robust to shape diversity.
Therefore, we introduce pre-training for the model encoder using a large-scale set of individual images.
Specifically, we employ Masked Autoencoder (MAE) \cite{He_2022_CVPR}, which can learn structural and shape features from unlabeled data, as the pre-training method.
Finally, we train the proposed segmentation model using the proposed PCI, utilizing the ViT backbone pre-trained with MAE.
We demonstrate the effectiveness of the proposed PCI and segmentation model through performance evaluation experiments using real plankton community images.

\section{Related Work}

Research applying deep learning-based image processing to plankton image analysis has been conducted to automate plankton monitoring.
While optical microscopes are relatively cost-effective and widely used for monitoring, the resulting images are typically ``crowded images'' containing multiple plankton individuals.
Therefore, realizing an automated monitoring system requires both the detection of individuals from crowded images and the classification of the detected individuals.
Regarding the automatic classification of extracted individual images, specialized models have been proposed leveraging large-scale public datasets.
On the other hand, research on individual detection from crowded images remains limited due to the scarcity of training data, with most studies restricted to applying general object detection models to small-scale, privately constructed datasets.

\subsection{Plankton Image Classification}

Large-scale datasets are indispensable for deep learning models to achieve high performance.
For classification tasks, several large-scale datasets of plankton individual images with image-level labels have been released \cite{BearingSea,WHOI,Gorsky,Kyathanahally-FM-2021,BNMNSS-B-2024-Otake}, and classification methods utilizing them have been proposed \cite{2018-LOM-Luo,Lumini1,Kyathanahally-FM-2021,Ito-ACPR-2023}.
For instance, Ito et al. proposed Hierarchical Attention Branch Network (H-ABN) \cite{Ito-ACPR-2023}, which extends ABN \cite{ABN}. 
Specifically, this method improves classification accuracy by hierarchically attending to discriminative regions based on biological taxonomic ranks (e.g., Order, Family, Genus).
The dataset used in the H-ABN experiments has been released as the FREPJ-Z dataset \cite{BNMNSS-B-2024-Otake}.
The FREPJ-Z dataset consists of 61,529 zooplankton individual images collected from lakes and dam reservoirs in Japan, with taxonomic labels assigned for ``Class,'' ``Order,'' ``Family,'' ``Genus,'' and ``Species.''
Although research on classifying cropped individual images has progressed, applying these models to actual monitoring requires accurate cropping of individuals from crowded images as a pre-processing step.

\subsection{Plankton Detection}

Crowded images captured by optical microscopes frequently contain debris mixed with plankton and overlapping individuals.
Simple thresholding or connected component analysis often fails in such scenarios, leading to false detections of debris or inability to separate overlapping individuals.
To address this issue, the application of instance segmentation, which identifies object regions at the pixel level, is being explored.
Bergum et al. \cite{Bergum_2020} employed Mask R-CNN \cite{He2017MaskR}, a major instance segmentation model, to detect copepods.
In their study, a dataset of 126 community images was created from collected water samples, with 776 copepod individuals annotated.
From this dataset, 88 images (containing 541 individuals) were used for training.
Similarly, Pana{\"i}otis et al. \cite{Panaotis2022ContentAwareSO} used Mask R-CNN to detect marine plankton using 106 images (3,356 individuals, 24 classes) captured by an In Situ Ichthyoplankton Imaging System (ISIIS) \cite{Cowen2008InSI}.

Two major challenges remain in plankton segmentation.
The first is the scarcity of training data.
All existing studies rely on small-scale private datasets created independently, and no large-scale public dataset exists.
The extremely high cost of pixel-level annotation for all individuals in community images hinders research progress.
The second challenge lies in the structural limitations of the models.
Most existing methods directly apply CNN-based models like Mask R-CNN \cite{He2017MaskR}.
Since CNNs prioritize local features, there is a concern that accuracy may degrade when separating densely overlapping objects or handling plankton with high shape diversity, as the model may fail to capture the global context.
In this paper, we resolve the data scarcity issue through PCI generation and improve the capability to handle overlaps and shape variations using a ViT-based model.

\section{Pseudo Community Image Generation}

\begin{figure*}[t]
  \centering
  \includegraphics[width=\linewidth]{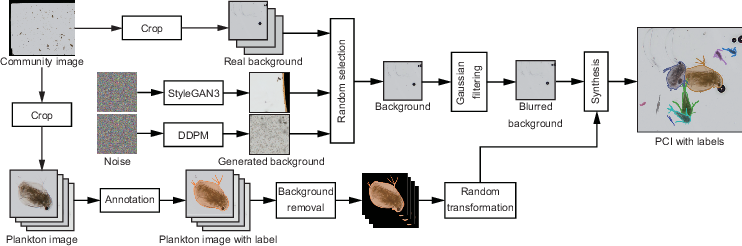}
  \caption{Overview of the Pseudo Community Image (PCI) generation process.}
  \label{fig:makingpci}
\end{figure*}

In this section, we describe the procedure for generating Pseudo Community Images (PCI), as illustrated in Fig. \ref{fig:makingpci}.
The proposed method generates PCI by synthesizing individual images onto background images derived from real crowded images.
The details of each process are described below.

\subsection{Background Images}

For the background images of PCI, we use both background regions extracted from real plankton crowded images and synthetic background images generated by generative models.
By extracting plankton-free regions from real community images and using them as backgrounds, we can train the model to accurately detect plankton even in environments containing non-plankton objects such as debris.
Furthermore, to expand background diversity, we also employ image generative models.
Specifically, we generate background images using StyleGAN3 \cite{NEURIPS2021_076ccd93} and Denoising Diffusion Probabilistic Models (DDPM) \cite{Ho2020DenoisingDP}.
Note that these generative models are fine-tuned using real background images.

\subsection{Individual Images}

For individual images used in synthesis, we utilize the publicly available FREPJ-Z dataset \cite{BNMNSS-B-2024-Otake}.
This dataset consists of individual plankton images labeled with taxonomic names: ``Class,'' ``Order,'' ``Family,'' ``Genus,'' and ``Species.''
For these images, we manually assigned pixel-level labels to the plankton regions and removed the background regions.

\subsection{PCI Creation}

The procedure for PCI creation is as follows.
We use images randomly selected from the aforementioned background and individual images.
First, to increase background variation, we apply random vertical and horizontal flips to the selected background image.
In actual microscopic photography, the background region is not always in focus.
To reproduce this situation, we apply a Gaussian filter to add a blur effect to the background image.
The standard deviation $\sigma$ of the Gaussian filter is randomly determined from the range $[0, 2)$ for each image.
Next, the number of individual images to be synthesized into a single PCI is randomly determined within the range of 6 to 10.
To expand the variation of plankton individuals, random flipping, rotation, and scaling are applied to the individual images.
Finally, multiple individual images are placed and synthesized onto the background image, allowing for overlaps between individuals and truncation at the image borders, thereby creating a natural community image.

\subsection{PCI Labeling}

Simultaneously with PCI creation, pixel-level ground truth labels for plankton individuals and background regions are automatically generated.
The same geometric transformations (flipping, rotation, scaling, and placement position) applied to the individual images are applied to the corresponding individual mask images.
By synthesizing these masks onto a background mask, segmentation labels corresponding to the PCI are generated.
At this stage, we consider the granularity of the class labels.
For plankton, shape differences between individuals become more minute as the taxonomic rank becomes lower (e.g., species level).
To enable the segmentation model to learn global shape features of plankton, we adopt ``Family'' level labels from the taxonomic labels provided in the FREPJ-Z dataset.
Regions other than plankton are labeled as the background class. Through the above procedure, we construct a labeled PCI dataset for supervised training.

\section{Plankton Image Segmentation Method}

\begin{figure*}[t]
  \centering
  \includegraphics[width=\linewidth]{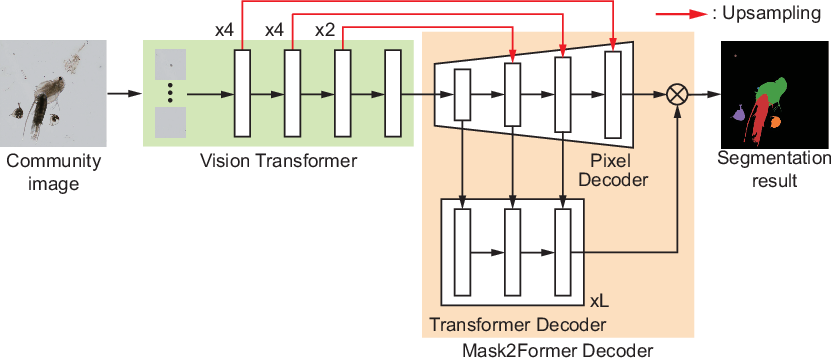}
  \caption{Overview of the plankton image segmentation method {\it PlankFormer}.}
  \label{fig:method}
\end{figure*}

Capturing global image features is essential to separate and detect plankton in community images where debris and overlapping individuals are significant.
In this paper, we propose {\it PlankFormer}, a plankton detection method based on global feature representation using an instance segmentation model that adopts a ViT \cite{Dosovitskiy-ICLR-2021} as the encoder.
An overview of the proposed method is shown in Fig. \ref{fig:method}. 
A key feature of the proposed method is the introduction of pre-training for the model encoder using a large-scale set of individual images.
This enables high generalization performance even when the segmentation model is fine-tuned using only PCI synthesized from a small number of labeled individual images.
The details of the proposed method are described below.

\subsection{Network Architecture}

The network architecture of the proposed method is designed based on Mask2Former \cite{Cheng2021MaskedattentionMT}.
We employ ViT-Large, consisting of 24 Transformer Encoder Blocks, as the encoder, and use the Pixel Decoder and Transformer Decoder of Mask2Former for the decoder.
Plankton appear in community images at various scales since individual sizes differ significantly depending on the species.
Therefore, the segmentation model requires the ability to detect objects across a wide range of sizes.
To perform high-precision detection based on multi-scale features, we utilize feature maps not only from the final layer of the encoder but also from intermediate layers as input to the decoder.
Specifically, we use the outputs from the 5th, 8th, and 16th layers of the Transformer Encoder Blocks.
Since the output of ViT is a sequence of patch tokens, we rearrange them to correspond to their spatial arrangement and convert them into 2D feature maps.
Since the ViT feature maps have a resolution of 1/32 of the input image in the experiments, we perform resolution conversion before inputting them to the decoder.
We apply upsampling to the feature maps of each extracted layer to construct a feature pyramid with resolutions of 1/8, 1/16, and 1/32 of the input image size.
The decoder uses these multi-scale features to output the plankton regions and class labels in the image.

\subsection{Training}

To achieve high generalization performance, deep learning-based segmentation models require a large amount of training data with diverse variations.
However, since the proposed PCI is generated by repeatedly synthesizing a small number of individual images, the diversity of individuals is inherently lower compared to actual community images.
Training a model from scratch using only PCI may cause the model to overfit to the biased features specific to PCI, leading to a failure in adapting to shape changes and overlaps in real images, which can degrade detection accuracy.
To address this issue, we introduce self-supervised pre-training using MAE \cite{He_2022_CVPR}.
Self-supervised learning allows training using only unlabeled images, enabling the utilization of a large number of target individual images without annotation costs.
MAE masks a large portion of the input image and trains the model to reconstruct the missing pixels from the remaining visible parts.
Through this task of inferring the whole from parts, the model acquires structural features of plankton and the ability to complete unseen parts.
This capability is particularly crucial for recognizing plankton in community images where debris and occlusions occur frequently, as the model needs to infer the shape of partially hidden bodies.
In the pre-training phase, we train the encoder using individual images from the FREPJ-Z dataset.
After pre-training, we combine the pre-trained encoder with an initialized decoder and fine-tune the entire model using labeled PCI.
During fine-tuning, we do not freeze the encoder weights, that is, we update all layers to adapt the model to the segmentation task.

\section{Experiments and Discussion}

In this section, we present the experiments conducted to evaluate the effectiveness of the proposed method.

\subsection{Dataset}

In this experiment, we utilized the FREPJ-Z dataset \cite{BNMNSS-B-2024-Otake}, constructed from optical microscope images of zooplankton samples collected from lakes in Japan.
The individual images contained in the FREPJ-Z dataset were used for pre-training the ViT \cite{Dosovitskiy-ICLR-2021}, generating PCIs, and training comparative methods.
For pre-training, we used 9,712 images containing individuals belonging to Cladocera, Copepoda, or Rotifera, which are the major zooplankton groups inhabiting Japanese lakes.

To generate background images for PCI generation, we employed generative models, specifically StyleGAN3 \cite{NEURIPS2021_076ccd93} and Denoising Diffusion Probabilistic Models (DDPM) \cite{Ho2020DenoisingDP}.
We trained these models using 60 real background images extracted from crowded images that were not used for evaluation.
For StyleGAN3 training, we set the batch size to 32 and the number of epochs to 100.
For DDPM training, the batch size was set to 4 and the number of epochs to 500.
Using each trained model, we generated 60 images each, serving as background images for PCI creation.
Additionally, as individual images for PCI generation, we used 160 images (16 families, 10 images each) from the FREPJ-Z dataset.
Individuals in these images also belong to Cladocera, Copepoda, or Rotifera.
Using these individual and background images, we generated a total of 4,800 PCIs to train the image segmentation model.
The total number of plankton individuals contained in the training PCIs is 33,851.
Details of the generated PCI dataset are listed in Table \ref{tbl:dataset}.

\begin{figure*}[t]
  \centering
  \includegraphics[width=.9\linewidth]{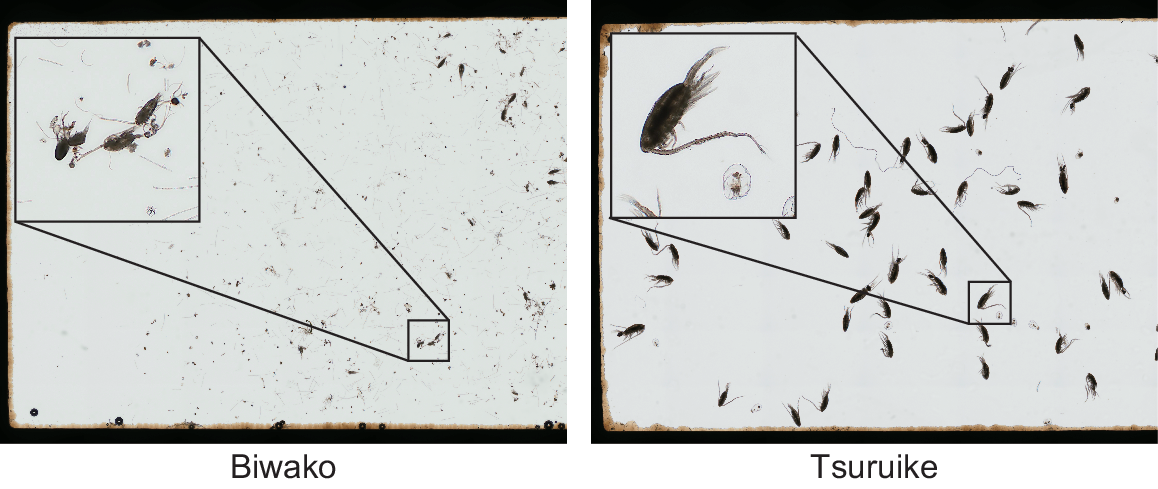}
  \caption{Evaluation datasets used in the experiments.}
  \label{fig:test_dataset}
\end{figure*}

For the evaluation of the image segmentation model, we used real community images included in the FREPJ-Z dataset.
Pixel-level labels were manually annotated for all plankton individuals belonging to Cladocera, Copepoda, and Rotifera in the community images used for evaluation.
In this experiment, to evaluate model performance under different environmental conditions, we employed two evaluation datasets: ``Biwako'' and ``Tsuruike.''
The Biwako dataset is a community image containing many relatively small individuals and a significant amount of debris.
On the other hand, the Tsuruike dataset is a clear community image containing relatively large individuals with little debris.
The number of target individuals for evaluation in each dataset is 95 for the Biwako dataset and 63 for the Tsuruike dataset.
The imaging area is $25 \times 20$ mm with a magnification of $40\times$, and the image resolution is approximately $10,000 \times 8,000$ pixels.
As input to the model, we used images cropped into $1,000 \times 1,000$ pixel patches.
Note that these patches were cropped with an overlap of 200 pixels between adjacent regions.
Detection results for each patch were integrated to perform evaluation at the original image size.
Evaluation was conducted separately for each dataset.
Fig. \ref{fig:test_dataset} shows examples of crowded images from the Biwako and Tsuruike datasets.

\begin{table}[t]
  \caption{Details of the generated PCI dataset, including the number of source individual and background images.}
  \label{tbl:dataset}
  \centering
  \begin{tabular*}{\linewidth}{@{\extracolsep{\fill}}cccccc}
    \toprule
    \multicolumn{2}{c}{Individual images} & \multicolumn{2}{c}{Background images} & \multicolumn{2}{c}{PCI} \\
    \cmidrule(lr){1-2} \cmidrule(lr){3-4} \cmidrule(lr){5-6}
    \# of images & \# of classes & Real & Generated & Images & Individuals \\
    \midrule
    160 & 16 & 60 & 120 & 4,800 & 33,851 \\
    \bottomrule
  \end{tabular*}
\end{table}

\subsection{Experimental Conditions}

In this experiment, to demonstrate the effectiveness of the proposed method for plankton detection, we compare detection accuracy with conventional instance segmentation methods: Mask R-CNN \cite{He2017MaskR} and Mask2Former \cite{Cheng2021MaskedattentionMT}.
To verify the effectiveness of pre-training using MAE \cite{He_2022_CVPR} in the proposed method, we conduct comparative experiments under the following conditions: (i) no pre-training, (ii) pre-training using the UrFound framework \cite{Yu2024UrFoundTU}, (iii) pre-training using the MoCo framework \cite{Chen2021AnES}, and (iv) pre-training via an individual image classification task (Family-level classification).
Furthermore, to verify the effectiveness of the PCI generated by the proposed method, we conduct an ablation study on the PCI generation method.
Specifically, we compare accuracy changes based on the taxonomic rank assigned to PCI labels (Order vs. Class), the type of background images (presence or absence of generated backgrounds), and the presence or absence of blur processing on the background.
For the encoder of the proposed method, we use ViT-Large \cite{Dosovitskiy-ICLR-2021}.
The input image size is set to $384 \times 384$ pixels, and the patch size is 32.
For the encoders of Mask R-CNN and Mask2Former, we use ResNet-50 \cite{ResNet}.

In the pre-training of the encoder, individual plankton images are used.
For pre-training via the classification task, the Family-level taxonomic group assigned to the individual images is used as the ground truth label.
For pre-training using MAE, we use AdamW \cite{AdamW} as the optimizer with an initial learning rate of 0.00025.
As data augmentation for pre-training, we apply horizontal and vertical flipping, random cropping (scale $0.6 \sim 1.4$, aspect ratio $0.8 \sim 1.2$), and color jittering (random changes in brightness, contrast, saturation, and hue).
The batch size is set to 64, and the number of epochs is 400.

For training the segmentation model (fine-tuning), the generated PCIs are used.
We use Detectron2\footnote{\url{https://github.com/facebookresearch/detectron2}} for the implementation.
AdamW \cite{AdamW} is used as the optimizer with an initial learning rate of 0.0001, employing a step learning rate schedule with a weight decay of 0.05.
As data augmentation during training, we apply Large-Scale Jittering (LSJ) \cite{Ghiasi2020SimpleCI} and random horizontal flipping.
The range of LSJ is set to $[0.1, 2.0]$.
Note that when pre-training with individual plankton images is not performed, weights pre-trained on ImageNet \cite{Deng2009ImageNetAL} are used as the initialization.
For Mask R-CNN and Mask2Former, we fine-tune models pre-trained on the COCO dataset \cite{Lin2014MicrosoftCC} using PCIs.
In all experiments, the number of iterations for fine-tuning the segmentation model is set to 30,000.

\subsection{Evaluation Metrics}

We evaluate the class-agnostic segmentation accuracy, focusing solely on detecting plankton as foreground objects rather than classifying their species.
We employ standard metrics from the COCO dataset \cite{Lin2014MicrosoftCC}: Average Precision (AP), specifically reporting mAP, AP$_{50}$, and size-dependent APs (AP$_S$, AP$_M$, AP$_L$).
AP is calculated based on the Intersection over Union (IoU) between the prediction $p$ and ground truth $g$:
\begin{equation}
  \mathrm{IoU} = \frac{|p \cap g|}{|p \cup g|},
\end{equation}
where $|\cdot|$ denotes the pixel count.
AP$_{50}$ is calculated at an IoU threshold of 0.50.
mAP is the average AP over IoU thresholds $\tau$ ranging from 0.50 to 0.95 with a step of 0.05:
\begin{equation}
  \mathrm{mAP} = \frac{1}{10} \sum_{\tau \in {0.50, 0.55, \dots, 0.95}} \mathrm{AP}{\tau}.
\end{equation}
AP$_S$, AP$_M$, and AP$_L$ represent mAP for small ($<32^2$ pixels), medium ($32^2 \sim 96^2$ pixels), and large ($>96^2$ pixels) objects, respectively.

\begin{table}[t]
  \caption{Ablation study on the ``Biwako'' dataset. \textbf{Bold} indicates the best results, and \underline{underlined} indicates the second-best results.}
  \label{tbl:abulation_Biwako}
  \centering
  \begin{tabular*}{\linewidth}{@{\extracolsep{\fill}}cccccccc}
    \toprule
    Rank & Gen. BG & Blur & mAP$\uparrow$ & AP$_{50}\uparrow$ & AP$_{S}\uparrow$ & AP$_{M}\uparrow$ & AP$_{L}\uparrow$ \\
    \midrule
    \multirow{4}{*}{Family} & --- & --- & 0.0503 & 0.1221 & 0.0138 & 0.0606 & \textbf{0.3370} \\
     & \checkmark & --- & \underline{0.0694} & 0.1541 & \underline{0.0600} & 0.0713 & 0.2083 \\
     & --- & \checkmark & 0.0531 & 0.1374 & 0.0293 & 0.0610 & 0.2699 \\
     & \checkmark & \checkmark & \textbf{0.0738} & \textbf{0.1749} & 0.0536 & \textbf{0.0783} & 0.2340 \\
    \midrule
    Order & \checkmark & \checkmark & 0.0460 & 0.1179 & \textbf{0.0778} & 0.0279 & 0.1852 \\
    \midrule
    Class & \checkmark & \checkmark & 0.0675 & \underline{0.1649} & 0.0323 & \underline{0.0725} & \underline{0.2750} \\
    \bottomrule
  \end{tabular*}
\end{table}

\begin{table}[t]
  \caption{Ablation study on the ``Tsuruike'' dataset. \textbf{Bold} indicates the best results, and \underline{underlined} indicates the second-best results.}
  \label{tbl:abulation_Tsuruike}
  \centering
  \begin{tabular*}{\linewidth}{@{\extracolsep{\fill}}cccccccc}
    \toprule
    Rank & Gen. BG & Blur & mAP$\uparrow$ & AP$_{50}\uparrow$ & AP$_{S}\uparrow$ & AP$_{M}\uparrow$ & AP$_{L}\uparrow$ \\
    \midrule
    \multirow{4}{*}{Family} & --- & --- & \textbf{0.6580} & \textbf{0.9532} & --- & 0.5832 & \textbf{0.6597} \\
     & \checkmark & --- & 0.5801 & 0.8952 & --- & 0.5832 & 0.5898 \\
     & --- & \checkmark & 0.5692 & 0.8881 & --- & \textbf{0.7302} & 0.5650 \\
     & \checkmark & \checkmark & 0.5459 & 0.8893 & --- & \underline{0.7252} & 0.5407 \\
    \midrule
    Order & \checkmark & \checkmark & \underline{0.5946} & \underline{0.9154} & --- & 0.6515 & \underline{0.5960} \\
    \midrule
    Class & \checkmark & \checkmark & 0.5314 & 0.8572 & --- & 0.7010 & 0.5302 \\
    \bottomrule
  \end{tabular*}
\end{table}

\begin{table}[t]
  \caption{Experimental results on the ``Biwako'' dataset. \textbf{Bold} indicates the best results, and \underline{underlined} indicates the second-best results.}
  \label{tbl:quantity_biwako}
  \centering
  \begin{tabular*}{\linewidth}{@{\extracolsep{\fill}}llcccccc}
    \toprule
    Method & Backbone & Pretrain & mAP$\uparrow$ & AP$_{50}\uparrow$ & AP$_{S}\uparrow$ & AP$_{M}\uparrow$ & AP$_{L}\uparrow$ \\
    \midrule
    Mask R-CNN & ResNet-50 & --- & \underline{0.0538} & 0.0986 & \underline{0.0546} & \underline{0.0496} & 0.1065 \\
    Mask2Former & ResNet-50 & --- & 0.0170 & 0.0417 & 0.0040 & 0.0104 & 0.2287 \\ 
    \midrule
    Baseline & ViT-Large & --- & 0.0440 & 0.1045 & \textbf{0.0556} & 0.0277 & 0.2300 \\
    Baseline & ViT-Large & UrFound & 0.0442 & 0.1079 & 0.0016 & 0.0353 & \underline{0.3111} \\
    Baseline & ViT-Large & MoCo & 0.0436 & 0.1022 & 0.0125 & 0.0228 & \textbf{0.3136} \\
    Baseline & ViT-Large & Classification & 0.0364 & \underline{0.1109} & 0.0089 & 0.0293 & 0.1798 \\
    \midrule
    Proposed & ViT-Large & MAE & \textbf{0.0738} & \textbf{0.1749} & 0.0536 & \textbf{0.0783} & 0.2340 \\
    \bottomrule
  \end{tabular*}
\end{table}


\begin{table}[t]
  \caption{Experimental results on the ``Tsuruike'' dataset. \textbf{Bold} indicates the best results, and \underline{underlined} indicates the second-best results.}
  \label{tbl:quantity_Tsuruike}
  \centering  
  \begin{tabular*}{\linewidth}{@{\extracolsep{\fill}}llcccccc}
    \toprule
    Method & Backbone & Pretrain & mAP$\uparrow$ & AP$_{50}\uparrow$ & AP$_{S}\uparrow$ & AP$_{M}\uparrow$ & AP$_{L}\uparrow$ \\
    \midrule
    Mask R-CNN & ResNet-50 & --- & 0.3734 & 0.7739 & --- & 0.4126 & 0.3790 \\
    Mask2Former & ResNet-50 & --- & 0.5152 & 0.8037 & --- & \textbf{0.7515} & 0.5157 \\ 
    \midrule
    Baseline & ViT-Large & --- & \textbf{0.5926} & \textbf{0.9671} & --- & 0.4645 & \textbf{0.6004} \\
    Baseline & ViT-Large & UrFound & 0.4632 & 0.7681 & --- & 0.5195 & 0.4692 \\
    Baseline & ViT-Large & MoCo & 0.5438 & 0.8883 & --- & \underline{0.7505} & \underline{0.5446} \\
    Baseline & ViT-Large & Classification & 0.2970 & 0.6644 & --- & 0.6168 & 0.2902 \\
    \midrule
    Proposed & ViT-Large & MAE & \underline{0.5459} & \underline{0.8893} & --- & 0.7252 & 0.5407 \\
    \bottomrule
  \end{tabular*}
\end{table}

\begin{figure*}[t]
  \centering
  \includegraphics[width=\linewidth]{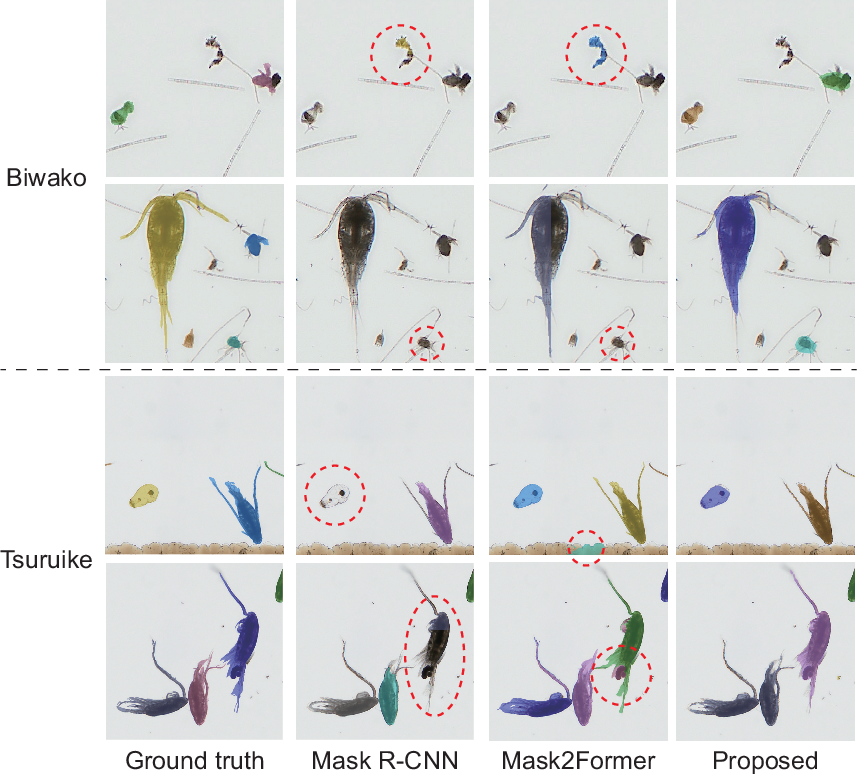}
  \caption{Example of segmentation results (zoomed-in views). Red dashed circles indicate false positives and false negatives.}
  \label{fig:result}
\end{figure*}

\subsection{Ablation Study}

We present the results of verifying the effectiveness of the PCI generation conditions in the proposed method in Table \ref{tbl:abulation_Biwako} and Table \ref{tbl:abulation_Tsuruike}.
Note that AP$_S$ is excluded from the evaluation for the Tsuruike dataset as it does not contain small-sized individuals.
First, we discuss the impact of label granularity.
For the Biwako dataset (Table \ref{tbl:abulation_Biwako}), using ``Family''-level labels yielded the highest mAP and AP$_{50}$.
The Biwako dataset contains many small individuals and frequent occlusions caused by debris.
In such complex environments, training with ``Family''-level labels, which properly reflect the fine-grained shape differences among individuals, likely contributed to the improvement in detection accuracy.
Conversely, for the Tsuruike dataset (Table \ref{tbl:abulation_Tsuruike}), under the conditions where generated backgrounds and blur processing were applied, using ``Order''-level labels resulted in the highest mAP and AP$_{50}$.
Since the Tsuruike dataset has simple backgrounds and high individual visibility, using coarser-grained labels allowed the model to learn more generalizable shape features, thereby suppressing overfitting.
Next, we examine the effects of background images and blur processing.
For the Biwako dataset (Table \ref{tbl:abulation_Biwako}), the proposed method, which combines generated backgrounds with Gaussian blur, achieved the highest accuracy.
In particular, AP$_M$ improved significantly compared to the case without generated backgrounds and blur.
In contrast, for the Tsuruike dataset in Table \ref{tbl:abulation_Tsuruike}), simple PCI without these augmentations yielded higher accuracy.
This suggests that because the Tsuruike images are clear, domain diversification techniques such as blurring inadvertently introduced a domain gap.
However, in practice, the system is required to handle challenging conditions such as debris and defocus, as seen in the Biwako dataset.
Therefore, considering robustness in real-world environments, expanding diversity through generated backgrounds and blur processing is essential.

\subsection{Experimental Results}

In this section, we verify the individual detection performance of the proposed method on community images and evaluate its effectiveness through comparisons with conventional methods.
First, we present the quantitative evaluation results for the Biwako dataset, which contains significant debris and small-sized individuals, in Table \ref{tbl:quantity_biwako}.
The proposed method achieved the highest performance in mAP, AP$_{50}$, and AP$_M$ among all comparative methods.
Notably, compared to the Baseline without pre-training, the proposed method with MAE pre-training significantly improved mAP.
It also demonstrated higher accuracy compared to other pre-training methods such as UrFound and MoCo.
Furthermore, the higher AP$_S$ compared to the CNN-based Mask2Former suggests that the ViT encoder successfully captures global context, thereby contributing to the detection of small individuals that are often buried in debris.
These results indicate that MAE pre-training is extremely effective for crowded images with complex backgrounds.
Next, we present the results for the Tsuruike dataset, which contains relatively little debris and large-sized individuals, in Table \ref{tbl:quantity_Tsuruike}.
Note that AP$_S$ is excluded from the evaluation as the Tsuruike dataset does not contain small individuals.
The proposed method achieved the second-best accuracy after the Baseline and outperformed conventional methods such as Mask2Former.
Comparing the Baseline and the proposed method, while AP$_M$ improved with MAE pre-training, AP$_L$ tended to decrease.
This trend was consistent across other pre-training methods.
It suggests that for large individuals (AP$_L$) with clear shapes that are easy to distinguish, domain-specific learning (on PCI) may have been more advantageous than the strong shape priors acquired through pre-training.
However, addressing adverse conditions like those in the Biwako dataset is crucial for real-world monitoring.
The proposed method demonstrates significant performance gains on the challenging dataset while maintaining stable accuracy on the easier dataset.
Fig. \ref{fig:result} shows qualitative examples of detection results by each method.
It can be confirmed that the proposed method exhibits fewer false positives (misidentifying debris) and false negatives (missing overlapping individuals) compared to conventional methods, accurately separating individuals into distinct masks.

These results demonstrate that the combination of MAE pre-training and PCI generation with expanded diversity is effective for high-precision individual detection from plankton community images under various imaging conditions.

\section{Conclusion}

In this paper, we proposed a robust plankton segmentation framework for automated monitoring.
To overcome the scarcity of labeled training data and handle complex occlusions in community images, we introduced a pipeline for generating labeled PCI using generative models and employed a ViT encoder pre-trained with MAE.
Evaluation on real-world datasets demonstrated that our method outperforms state-of-the-art baselines like Mask2Former, particularly in debris-heavy environments where MAE pre-training and PCI diversity significantly reduced false detections.
These results highlight the potential for automating plankton monitoring without incurring large annotation costs.
Future work includes bridging the domain gap observed in clear images containing large individuals and extending the framework to detailed multi-class classification.

%
%
\bibliographystyle{splncs04}
\bibliography{paper}

\end{document}